\documentclass{article} % For LaTeX2e
\usepackage{iclr2024_conference,times}

\usepackage{amsmath,amsfonts,bm}

\def\eqref#1{equation~\ref{#1}}
\def\1{\bm{1}}

\DeclareMathAlphabet{\mathsfit}{\encodingdefault}{\sfdefault}{m}{sl}
\SetMathAlphabet{\mathsfit}{bold}{\encodingdefault}{\sfdefault}{bx}{n}

\usepackage[utf8]{inputenc} % allow utf-8 input
\usepackage[T1]{fontenc}    % use 8-bit T1 fonts
\usepackage{hyperref}       % hyperlinks
\usepackage{url}            % simple URL typesetting
\usepackage{booktabs}       % professional-quality tables
\usepackage{amsfonts}       % blackboard math symbols
\usepackage{nicefrac}       % compact symbols for 1/2, etc.
\usepackage{microtype}      % microtypography
\usepackage{xcolor}         % colors

\usepackage{hyperref}
\usepackage{url}
\usepackage{soul}
\usepackage{booktabs}
\usepackage{adjustbox}
\usepackage{wrapfig}
\usepackage{multirow}
\usepackage{ulem}
\usepackage{paralist}
\usepackage{threeparttable}

\title{Two-stage LLM Fine-tuning with Less Specialization and More Generalization}

\iclrfinalcopy

\author{Yihan Wang\thanks{~~Work done while at Google.} \\ 
Department of Computer Science\\
UCLA\\
\texttt{wangyihan617@gmail.com} \\
\And
Si Si\\
Google \\
\texttt{sisidaisy@google.com} \\
\And
Daliang Li\\
Google \\
\texttt{daliangli@google.com} \\
\And
Michal Lukasik\\
Google \\
\texttt{mlukasik@google.com} \\
\And
Felix Yu\\
Google \\
\texttt{felixyu@google.com} \\
\And
Cho-Jui Hsieh\\
Department of Computer Science\\           
UCLA \\
\texttt{chohsieh@cs.ucla.edu} \\
\And
Inderjit S Dhillon\\
Google \\
\texttt{isd@google.com} \\
\And
Sanjiv Kumar\\
Google \\
\texttt{sanjivk@google.com} \\
}

\begin{document}

\maketitle

\begin{abstract}

Pretrained large language models (LLMs) are general purpose problem solvers applicable to a diverse set of tasks with prompts. They can be further improved towards a specific task by fine-tuning on a specialized dataset. However, fine-tuning usually makes the model narrowly specialized on this dataset with reduced general in-context learning performances, which is undesirable whenever the fine-tuned model needs to handle additional tasks where no fine-tuning data is available. 
In this work, we first demonstrate that fine-tuning on a single task indeed decreases LLMs' general in-context learning performance. We discover one important cause of such forgetting, format specialization, where the model overfits to the format of the fine-tuned task.
% and is unable to output anything beyond this format.
We further show that format specialization happens at the very beginning of fine-tuning. To solve this problem, we propose Prompt Tuning with MOdel Tuning (ProMoT), a simple yet effective two-stage fine-tuning framework that reduces format specialization and improves generalization.
% alleviates the forgetting of in-context learning abilitites and improves generalization. 
ProMoT offloads task-specific format learning into additional and removable parameters by first doing prompt tuning and then fine-tuning the model itself with this soft prompt attached. 
With experiments on several fine-tuning tasks and 8 in-context evaluation tasks, we show that ProMoT achieves comparable performance on fine-tuned tasks to standard fine-tuning, but with much less loss of in-context learning performances across a board range of  out-of-domain evaluation tasks. More importantly, ProMoT can even enhance generalization on in-context learning tasks that are semantically related to the fine-tuned task, e.g. ProMoT on En-Fr translation significantly improves performance on other language pairs, and ProMoT on NLI improves performance on summarization.
Experiments also show that ProMoT can improve the generalization performance of  multi-task training. 

\end{abstract}

\section{Introduction}
\label{sec:intro}
% \begin{itemize}
%     \item Using pretraining to improve the ability of LLMs can be limited due to the noisy pretraining datasets
%     \item To improve the pretrained model, we can use fine-tuning on small but high-quality datasets
%     \item forgetting is severe for in-context learning abilities during fine-tuning
%     \item Two goals: Improve the performance on fine-tuning dataset while alleviate the forgetting on in-context learning tasks; 
%     % Improve the few-shot performance on some related tasks during fine-tuning.
%     \item contributions
% \end{itemize}
Natural language processing (NLP) has recently been revolutionized by scaling up transformer based large language models (LLMs) together with large-scale pretraining \citep{transformer, bert, t5, gpt3, gopher, palm, megatron, touvron2023llama}. In addition to improved downstream performances, these pretrained LLMs can perform a broad array of unforeseen tasks when provided with a prompt. This in-context learning capability allows users to flexibly re-purpose LLMs for specific tasks with a minimum amount of supervised data, making it extremely convenient for fast prototyping and experimentation, especially in the low data regime.

However, even the largest and most advanced LLMs leave a lot to be improved. Grounding and eliminating hallucinations \citep{maynez-etal-2020-faithfulness}, reasoning and logical clarity \citep{LLMReasoningDeepMind}, mathematics \citep{gpt3, noorbakhsh2021pretrained} are just a few examples where LLMs still lag behind the best human performances, or in some cases, the fine-tuned performances of the same model. 

The most common practice to improve a pretrained model is to fine-tune it on a specialized task or several tasks.
However, fine-tuning on LLM usually causes over-specialization to the fine-tuning tasks, and harm the model's pre-existing generalization ability on unseen tasks via in-context learning.
As we show later, an mT5 model finetuned on a single task loses its few-shot performance on unseen tasks within one thousand steps of fine-tuning. When faced with hundreds of downstream tasks and even unknown tasks, we expect to have a single fine-tuned model that is both superior on supervised fine-tuned tasks and general unseen tasks. Thus, it becomes very important to develop new techniques for finetuning that prevent over-specialization of these fine-tuned models only to a few tasks.
% The over-specialization from fine-tuning limits gradually improving a general language model during fine-tuning.
\begin{table*}[h!]
    \centering
    \adjustbox{max width=0.80\linewidth}{
    \begin{tabular}{c|c}
    \toprule
          Ground-Truth Output& Mercedes' Lewis Hamilton took the outright \\&championship lead for the first time this season \\&with a dominant victory in the \textbf{Italian} Grand Prix.
 \\
 \midrule
         Pretrained mT5& Hamilton won the {\color{red}\textbf{British}} Grand Prix.\\
\hline
         Fine-tuned mT5 on RTE & {\color{red}\textbf{True}}\\
\hline
         Fine-tuned mT5 with ProMoT (Ours) on RTE & Lewis Hamilton won the {\color{blue}\textbf{Italian}} Grand Prix.
\\
    \bottomrule
    \end{tabular}
    }
    \caption{Output comparison of pretrained and fine-tuned mT5 models vs. fine-tuned with ProMoT on the RTE binary classification NLI dataset, performing in-context 1-shot summarization.}
    \label{tab:example}
\end{table*}

In this work, we discover that the loss of general in-context learning abilities during fine-tuning is, to a large extent, caused by format specialization, which makes model overfitting to the specific task format. For example, an mT5 \citep{xue2020mt5} model learns in the output space with only ``True'' and ``False'' if we fine-tune it on a binary classification dataset, losing its ability to flexibly generate different output styles according to the in-context prompts of other tasks. We show that format specialization tends to happen at the very beginning of fine-tuning, before the model fully learns the semantic content of the task.

Based on these observations, we propose a simple solution to alleviate format specialization: PROmpt Tuning with MOdel Tuning (ProMoT), which off-loads format learning to a small amount of task-specific parameters that are external to the model. 
ProMoT is a two-stage fine-tuning process. At the first stage, we freeze the pretrained model and tune a small set of additional parameters, where we find adding soft prompt before the input \citep{lester-etal-2021-power} is a good choice. At the second stage, we freeze the additional parameters and tune the main model.
Since format information is learned first, it mostly enters the small set of additional parameters. At inference time, we 
can decide whether to remove the additional parameters depending on whether the incoming task share the same format as the fine-tuned task.

% use the model with the tuned prompt if the inference task has the same format as the fine-tuned task, otherwise we remove the tuned prompt and use the model directly. 
Our experiments show that ProMoT significantly alleviates specialization during fine-tuning, while boosting generalization on semantically related tasks with different formats. 
For example, fine-tuning the model only on an NLI binary classification dataset, a mT5 XXL model consistently obtains improved in-context learning performance on summarization compared with the pretrained model, possibly due to improved grounding learned from NLI. See Table \ref{tab:example} for a concrete example.  
With ProMoT, we can obtain models with both better supervised performance compared to pretrained models and better general in-context learning performance compared to standard finetuning.

To summarize, our contributions are 4-fold: 
% \michal{To save space, I think it's ok to omit itemization.}\daliangli{I kinda like the itemization. We can move 5.4 to the appendix before this becomes necessary. }
\begin{compactitem}
   \item We show empirically that general in-context learning capabilities decrease during single-task fine-tuning for T5 models. We identify format specialization as one of the important causes which mostly happens at the beginning of fine-tuning. 
    \item  We propose a novel 2-stage fine-tuning framework: PROmpt Tuning with MOdel Tuning (ProMoT) to reduce format specialization during fine-tuning. 
    % thus opens up opportunity to generalize on unseen in-context tasks.
    % \item We evaluate ProMoT on 10 different NLP tasks including classification, summarization, translation and question answering, and compare it with both the pretrained and traditionally fine-tuned models.
    \item Experiments on 10+ NLP tasks show that ProMoT significantly reduces specialization of fine-tuned models compared to standard fine-tuning, while reaching similar supervised performance. The reduction in specialization opens up opportunities to enhance generalization across very dissimilar tasks when they share some semantic aspects.
    % ; 3) When fine-tuned on a single task, ProMoT generalize across very dissimilar tasks when they share some semantic aspects; 
    % 4) When used in a multi-task setting, ProMoT result in better generalization to other in-context learning tasks compared to vanilla multi-task fine-tuning.
    
    \item ProMoT can be combined with many existing fine-tuning and parameter-efficient fine-tuning methods. We show examples where ProMoT is combined with multi-task fine-tuning and fine-tuning with 1-shot prompts to further boost the generalization on unseen tasks.

\end{compactitem}

\vspace{-5pt}
\section{Related Work}
\vspace{-5pt}
% \begin{itemize}
%     \item Large language models and in-context learning: T5, mT5, GPT, PALM...
%     \item Catastrophic forgetting 
%     \item prompt tuning: prompt tuning, multi-layer prompt tuning...
%     \item Transferbility: Meta-ICL, FLAN \citep{wei2021fine-tuned}, translation
% \end{itemize}
Pretrained LLMs are general problem solvers with in-context prompts \citep{raffel2020exploring, xue2020mt5, radford2018improving, palm, min2022rethinking, touvron2023llama}. \citet{zhai2023investigating} evaluates the catastrophic forgetting in multimodal language model fine-tuning, which is limited to image classification tasks. \citet{DataDistributionalProperties22, gao2020making} study the effect of pretraining data distribution on in-context learning on image recognition tasks, where the tension between in-context learning tasks and fine-tuning tasks is discussed. They propose changing the data distribution to ease such tension, which could be difficult for generative NLP tasks. ProMoT is an orthogonal method that does not require changes in data distribution. 

In a recent study, \citet{ramasesh2022effect} found that as model size increases, the model becomes less prone to catastrophic forgetting. 
% Catastrophic forgetting is the phenomenon where the knowledge acquired by a model in pretraining is forgotten during fine-tuning.
% This paper points out the increasing potential for larger models to mitigate forgetting. 
However such studies are mostly focused on tasks of similar format, e.g. a sequence of different classification tasks.
In this work we explore vastly different tasks, e.g. classification v.s. long form generation where the format itself is critical.

Different from full fine-tuning, prompt-tuning \citep{lester-etal-2021-power, zhang2021differentiable}, adapters and LoRA \citep{hu2021lora, he2021towards, houlsby2019parameter} adapt a pretrained model to a task with a small set of tunable parameters. Parameter-efficient methods like these largely leave the pretrained model intact, which can preserve the pre-existing in-context learning abilities. However, they also miss the opportunity to further improve the pretrained model with a small, high quality dataset that generalizes beyond the fine-tuned task. Besides, these parameter-efficient methods also underperform fine-tuning on the supervised task in many cases, as shown in \citep{lester-etal-2021-power, liu2021p} and in our results. 

% Prompt tuning \citep{lester-etal-2021-power, zhang2021differentiable} appends continuous trainable embeddings (soft prompt) before the inputs and optimizes the prompt.
% %Prompt-tuning doesn't modify the pretrained model and thus avoids catastrophic forgetting on in-context learning abilities. 
% Prompt tuning underperforms fine-tuning in many cases, as shown in \citep{lester-etal-2021-power, liu2021p} and in our results. Similarly, one can tune a small set of additional parameters with adapters or low rank matrices \citep{hu2021lora, he2021towards, houlsby2019parameter} to adapt a pretrained model to a task. Parameter-efficient methods like these largely leave the pretrained model intact. Thus they miss the opportunity to further improves the LLM with a small, high quality dataset that generalizes beyond the fine-tuned task.
% \michal{may be worth to cite more recent works, e.g. on adapters}

Another line of work uses multi-task fine-tuning to improve generalization on unseen in-context learning tasks.
% teach the model in-context few-shot or zero-shot abilities, providing a meta learning approach to in-context learning.
\cite{wei2021fine-tuned, chung2022scaling} fine-tune PaLM and T5 on large-scale multitask datasets with diverse natural language prompts, improving the zero- and few-shot performance on unseen tasks. 
\citet{min2021metaicl} incorporate the in-context learning objective into fine-tuning on multitask datasets with few-shot prompts.
This approach relies on multi-task training to generalize, while orthogonally, ProMoT improves the generalization of each single fine-tuning task, whether used in a multi-task setting or not. ProMoT can indeed be combined with multi-task training to obtain better generalization as we demonstrate in Sec.~\ref{sec:SequentialTrainingAndMultitaskTraining}.
% Such methods require fine-tuning on a large collection of different tasks to generalize, with each task being one datapoint of meta-learning. In contrast, ProMoT is not limited to this regime and shows positive task transfer with a single fine-tuning task. 
In addition, such approaches often require human engineered instructions or prompts for each task to partly alleviate format specialization, while ProMoT uses prompt tuning, which has two advantages: 1) ProMoT does not require the elaborate trial and error of prompt engineering as it optimizes the soft prompts with data. 2) Soft prompts are more effective at absorbing the format compared to natural language prompts, as shown in Table ~\ref{tab:ablation}.
% Finally, ProMoT is largely orthogonal to meta-learning style methods. They can be combined by applying ProMoT to a massively multitask setting where further improved generalization is expected. One could also apply ProMoT directly to the meta task of learning to solve in-context tasks. This, however, is beyond the scope of our current paper and will be left for future exploration.

\vspace{-5pt}
\section{Format specialization in fine-tuning causes the loss of in-context learning capabilities}
\label{sec:observation}
\vspace{-5pt}
In this section, we first show empirically with an mT5 XXL model that 1) in-context learning abilities are lost during fine-tuning; 2) format specialization is an important cause for such loss; 3) format specialization happens at the very beginning of fine-tuning. 
\subsection{Loss of in-context learning capabilities during fine-tuning} 
\label{sec:forget}

% \begin{itemize}
%     \item LLMs forgets in-context learning abilities quickly during fine-tuning
%     \item LLMs forgets by overfitting the data format in fine-tuning dataset
%     \item Show the curve when fine-tuning on RTE
%     \item Show that all the outputs are True/False on few-shot evaluation tasks: a table for different training steps
% \end{itemize}
%As indicated in \citet{DataDistributionalProperties22}, in-context learning performance of transformers can drop with supervised fine-tuning. 

% pretrained LLMs achieve good performance with in-context learning. However, in many applications one may want to fine-tune LLMs. 
In this subsection, we first show that the in-context learning performance usually drops significantly after standard fine-tuning. 

% \michal{A suggestion for organization of this section: how about we organize it in the way: research question -> how we are going to test it -> what we expect -> what we actually find. This way it would have a more 'scientific' feel.}

%This phenomenon was \cho{briefly?} discussed by \citet{DataDistributionalProperties22}, and 
%here we demonstrate this forgetting behavior with our training configurations on large language models.  \cho{Maybe say we conduct a more comprehensive experiment on multiple in-context/fine-tue tasks on a large language model? Otherwise it sounds like this has already been studied and not clear why we need to conduct experiments to verify this. }
In our experiments, we fine-tune a pretrained mT5 XXL model (13B parameters) \citep{xue2020mt5} on the Recognizing Textual Entailment (RTE)  dataset~\citep{wang2019superglue}. In RTE tasks, the model is required to predict ``True'' or ``False'' for whether the two given sentences are entailed. We fine-tune the mT5 model with default hyper-parameters and input/output template used in PaLM \citep{palm}. 

\begin{figure}[ht]
    \centering
    \begin{minipage}[t]{.42\textwidth}
        \centering
        \includegraphics[ width=\linewidth]{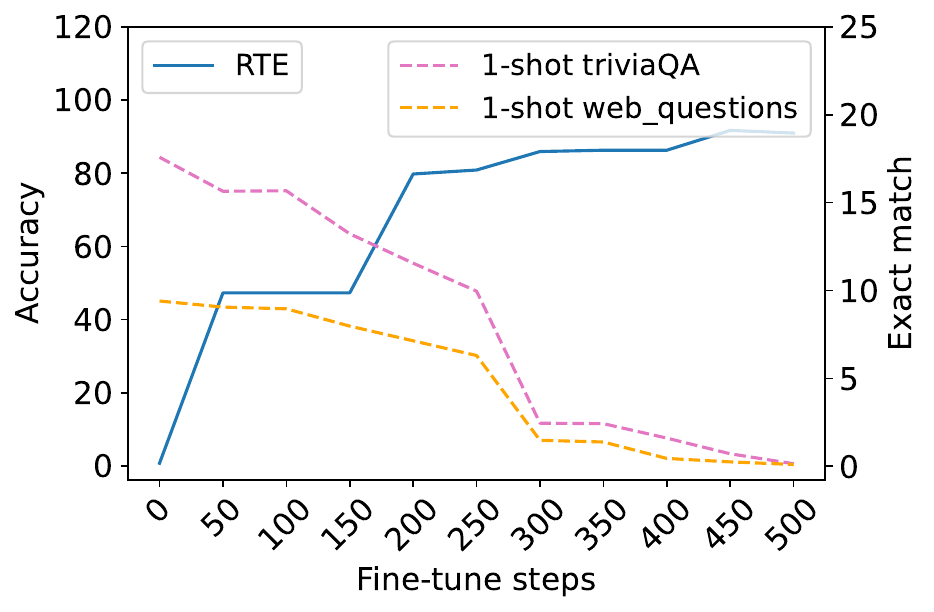}
        \vspace{-15pt}
        \caption{Loss of in-context learning abilities during fine-tuning. We show the learning curve of a model being fine-tuned on RTE dataset while being tested on 1-shot QA datasets. Left axis: Accuracy on RTE. Right axis: Exact match rate on 1-shot QA.
        }
        \label{fig:rte_qa}
    \end{minipage}%
    \hspace{0.05\textwidth}
    \begin{minipage}[t]{0.42\textwidth}
        \centering
        \includegraphics[width=\linewidth]{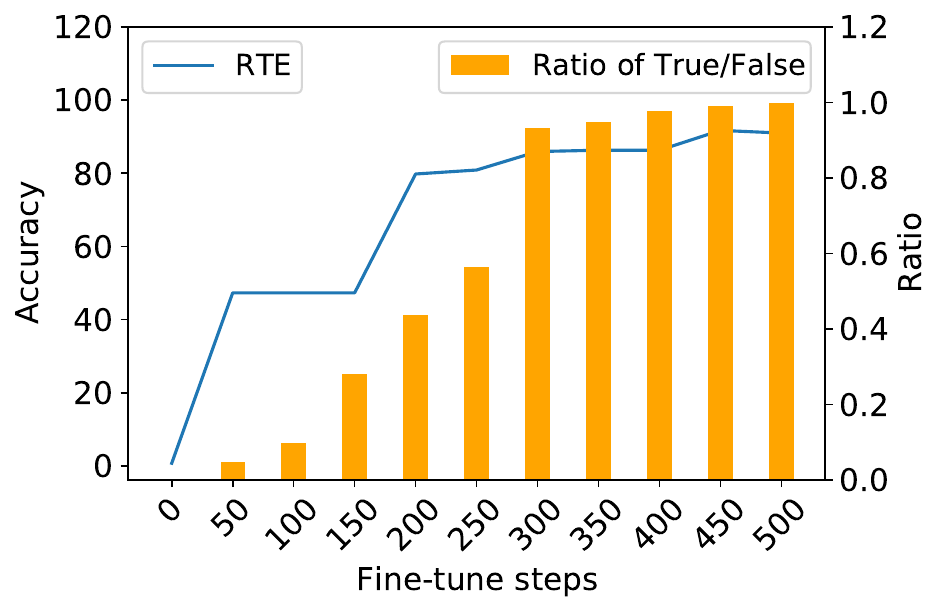}
 \vspace{-10pt}
        \caption{
        Format specialization in fine-tuning: showing the frequency of "True/False" style outputs when evaluated on 1-shot TriviaQA. The model is being fine-tuned on RTE. Left axis: Accuracy on RTE. Right axis: Ratio of True/False. 
        }
        \label{fig:true_false}
    \end{minipage}
    \vspace{-5pt}
\end{figure}

We want to see whether the model lost its in-context learning abilities on unseen task during fine-tuning. Therefore, we evaluate the fine-tuned model with two 1-shot QA tasks, TriviaQA \citep{joshi2017triviaqa} and web\_questions \citep{berant2013semantic}. The results are illustrated  in Figure \ref{fig:rte_qa}, where we can see that when the accuracy on RTE dataset increases with fine-tuning, performance on few-shot QA tasks drops drastically. 
This phenomenon is general and not a result of specific fine-tuning or evaluation tasks (more results in Section \ref{sec:exp_single}).

% \vspace{-5pt}
\subsection{Format specialization}
\label{sec:format}
Why are the in-context learning abilities of an LLM so easily lost after a few hundred steps of fine-tuning? A natural hypothesis is that due to the homogeneity of output formats in fine-tuning datasets, the model quickly specializes to this task format and learns to follow it no matter what the input sequence is. This leads to the loss of in-context learning abilities on other tasks that do not share the same format. here by ``format'' we refer to the common characteristics of the sequences in fine-tuning task as a subset of all possible sequences, such as the language used, typical input/output lengths and styles, special tokens or punctuation, upper/lower case styles etc. For example, the output format of RTE is a set of two labels, ``True'' or ``False'', among all possible sequences of tokens of various lengths. Since all data points share the same format in single-task fine-tuning, the model receives a strong gradient signal that the output should follow this format, thus its in-context learning performance on other tasks with different formats will drop, even when they share important semantic similarities with the fine-tuned task. 

To verify this hypothesis, we evaluate the RTE fine-tuned mT5 model on 1-shot TriviaQA task and count the percentage of outputs which are ``True'' or ``False''. Figure \ref{fig:true_false} shows that as the fine-tuning proceeds, the model outputs more ``True'' or `False' even with a 1-shot prompted input from TriviaQA. In particular, after 300 fine-tuning steps, $90\%$ of the output becomes ``True'' or ``False''. The same phenomenon happens on other in-context learning tasks. With a 1-shot WMT16 En-De translation prompt, after 500 steps of RTE fine-tuning, more than $99\%$ of the output becomes ``True'' or ``False''. 
This indicates that format specialization is a possible reason for the loss of general in-context learning capabilities during fine-tuning.

% \michal{this doesn't definitively prove that it's the format specialization behind only True/False outputs, right? For example, the model may have also lost the knowledge necessary to answer other tasks. Perhaps this claim could be softened.}

% \cho{I think we need to show results for multiple fine-tune and multiple in-context learning tasks (as Figure 1), to demonstrate this is not a special case. }

% Concluded from these observations, format plays an important role in the severe forgetting during fine-tuning. Different tasks can have different input/output formats which can be difficult to transfer between tasks, e.g. between binary classification and QA tasks. Since fine-tuning tends to over-fit the format for a specific task, it can significantly hurt the performance of most in-context learning tasks.  

% \begin{itemize}
%     \item Format is a stronger signal and can be learnt first
%     \item The gradient analysis
% \end{itemize}

\vspace{-5pt}
\subsection{Format learning happens first during standard fine-tuning}
\label{sec:format2}
\vspace{-5pt}
\begin{wrapfigure}{R}{0.4\linewidth}
  \centering
    \includegraphics[width=\linewidth]{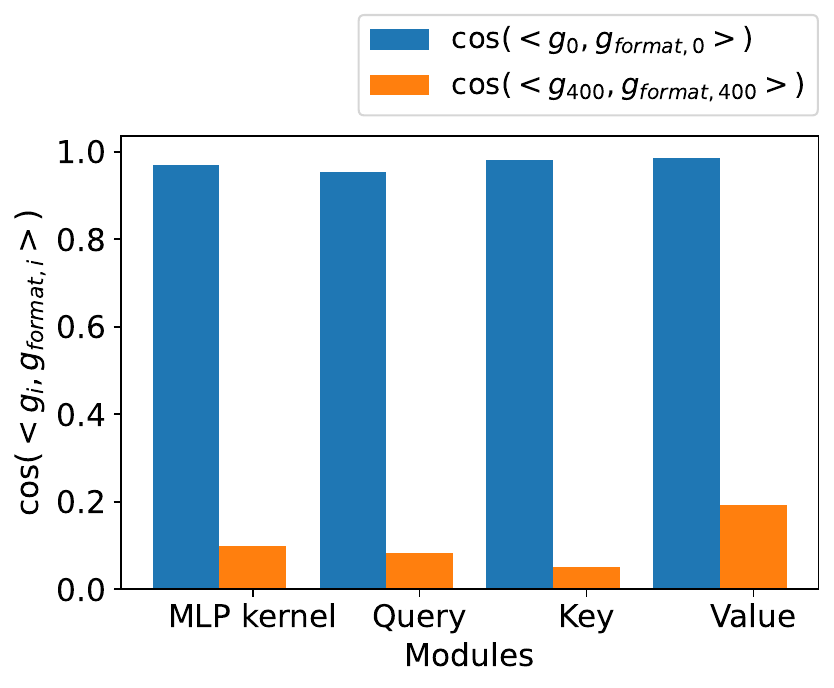}
    \caption{Format specialization happens at the beginning of fine-tuning: we show the cosine similarity between the full gradient $g$ and the format gradient $g_\text{format}$ on the MLP kernel, Query, Key and Value on the attention module. The $g$ and $g_{\text{format}}$ are much better aligned at the start of training, compared to at 400 steps. Comparison between more steps can be found in Figure \ref{fig:more_step_gradient} (Appendix).
%   \michal{Just to check: is it ok to have such floating figure in ICLR format? I haven't seen that before in an ICLR paper.}\daliangli{can we make this figure smaller (while keeping the fonts the same size)? }
%   \daliangli{can we re-order the bars to be say "Value, Query, Key, MLP kernel"? Otherwise the most visually attention seeking feature of this plot is the monotonic increase of the orange bars. Also, can we move the lengends a bit lower so people can see the top of the blue bars?}
% \daliangli{can we add a horizontal line at y=1?}
  }
  \label{fig:gradient}
  \vspace{-5pt}
\end{wrapfigure}  

Next, we show experimental evidence that format learning happens first during standard fine-tuning. This is not surprising as the overwhelming majority of fine-tuning data points have very similar formats, causing a gradient signal that dominates over others, more nuanced elements such as the semantic content of the task. 

More concretely, for the RTE dataset, the ``format'' refers to the fact that the $\text{output} \in \{\text{True}, \text{False}\}$, while the semantic content refers to the correlation between the input sequence and the output label. We isolate format learning from semantic learning by creating a randomized RTE dataset where the output labels are randomly shuffled, thus are no longer correlated with the input sequences. 
The gradients of format learning, $g_{\text{format}}$, are then given by the gradients on the randomized RTE dataset. By comparing with the full gradient $g$ on the original RTE we can detect when format learning happens during fine-tuning. We compute the gradients on the same batches of inputs for the two different settings.
Figure~\ref{fig:gradient} and Figure~\ref{fig:more_step_gradient} in Appendix show that at the very beginning of fine-tuning~(step 0), the full gradient $g$ is highly aligned with the format-only gradient $g_{\text{format}}$, signified by  $\cos(\langle g_0, g_\text{format,0} \rangle) \approx 1$. 
Since randomized RTE and original RTE share the format information only and contain totally different semantic content, this alignment implies that the model is mostly learning the format. 
After 400 fine-tuning steps, this alignment disappears where the cosine similarity drops to around $0.2$\footnote{We compute the format gradient at 400 steps, $g_{\text{format},400}$, by first fine-tuning the model on RTE for 400 steps, then computing the gradient on the randomized RTE dataset with the same batch of input sequences.}, when the True/False ratio reaches nearly 100\%.

\vspace{-5pt}
\section{Proposed Method: PROmpt  tuning  with MOdel Tuning (ProMoT)}
\label{sec:method}
\begin{figure}
    \centering
    \includegraphics[clip, trim=1.1cm 6.8cm 1.9cm 0.6cm, width=0.9\linewidth]{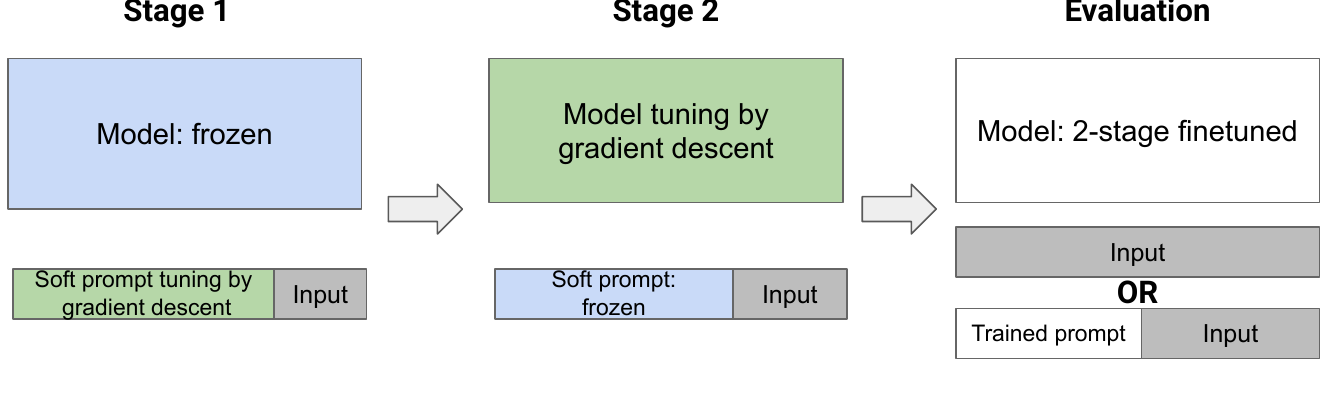}
    \vspace{-25pt}
    \caption{Overview of ProMoT, our two-stage fine-tuning strategy. We run prompt tuning at Stage 1 and model fine-tuning with the trained prompt at Stage 2. Green denotes trainable parameters and blue means frozen.
    % \michal{Would be nice to briefly discuss the method here.}
    }
    \label{fig:overview}
\end{figure}

% \begin{itemize}
%     \item Two-stage fine-tuning
%     \item Learning most of the format at the prompt-tuning stage with a small number of parameters
%     \item Train the pretrained model after capturing the format in learnable prompts
%     \item Can be combined with other methods like few-shot prompted training
% \end{itemize}
The observations from Section \ref{sec:observation} inspire us to decouple format learning from fine-tuning, in order to alleviate specialization to the fine-tuned task and preserve general in-context learning abilities. The key idea is to offload format learning to a separate small set of parameters during early fine-tuning, and allow the model's own parameter changes afterwards to focus more on the semantic content of the task. We propose a two-stage fine-tuning strategy called ProMoT, illustrated in Figure \ref{fig:overview}. At the first stage, ProMoT uses prompt tuning to capture the format in a trainable soft prompt while the model itself is frozen. At the second stage, ProMoT freezes the learned soft prompt and fine-tunes the model itself to focus on semantic skills that might be more transferable.
% For example, we can use few-shot fine-tuning at the second stage where the inputs are appended with few-shot prompts, which can further improve the few-shot performance as proposed by \citet{min2021metaicl}.

\paragraph{Stage 1: Prompt Tuning.}
Here we use a continuous trainable prompt (soft prompt)~\citep{lester-etal-2021-power} prepended before the embedded inputs as the separate small set of tunable parameters.
% Instead of updating the weight parameters, prompt tuning \citep{lester-etal-2021-power} prepends a continuous trainable prompt (soft prompt) before the embedded inputs and then updates this soft prompt during training. 
% The trainable prompt is a small set of free parameters taking the form of a few trainable embeddings. 
The soft prompt for a given fine-tuned task $P_e \in \mathbb{R}^{p\times e}$  is a small set of free parameters taking the form of a few trainable embeddings, where $p$ is the prompt length and $e$ is the embedding size.  
Given an input sequence, prompt tuning first embeds it with the text embedding layer of the pretrained model, and then prepends it with the soft trainable prompt. 
The soft prompt is then optimized to reduce the loss while the pretrained model is frozen.  
As indicated in Section \ref{sec:format2}, fine-tuning first learns the format. We expect that by  prompt tuning first, the soft prompt will learn the format. 
Although it is not guaranteed that the soft prompt only learns the format, the small capacity can prevent the soft prompt from learning all semantic skills in most realistic NLP tasks, as demonstrated by the performance gap between prompt tuning and standard fine-tuning.

% By running prompt-tuning at the first stage, 
% since fine-tuning learns format first, we suspect when replacing the first stage by prompt learning it will also try to use prompt to capture format information.  
% the format is learned first into the prompt, as indicated in Section \ref{sec:format} \
% \cho{I think we can't make this claim very strong since Section 4 is for standard fine-tuning, there's possibility that the situlation will be different if we update prompt instead of model. I think we can only say that since fine-tuning learns format first, we suspect when replacing the first stage by prompt learning it will also try to use prompt to capture format information.  }

% Formally, we have a soft prompt $P_e \in \mathbb{R}^{p\times e}$ for a given fine-tune task where $p$ is the prompt length and $e$ is the embedding size. Given an input sequence from the task $(x_1, x_2, ..., x_n)$, prompt-tuning first embeds the input sequence with the embedding layer in the pretrained model, then concatenates the soft prompt before the embeded input. The we optimize the soft prompt to reduce the loss while keeping the pretrained model is frozen. 

\paragraph{Stage 2: Fine-tuning with trained prompt.}
After prompt-tuning, we expect the trained prompt now storing most of the format information. We then freeze the soft prompt and fine-tune the pretrained model. Importantly, as shown in Figure \ref{fig:overview}, the soft prompt is still prepended before the input during this stage, forcing the model to learn things not captured already by the soft prompt. 

\paragraph{Other parameter-efficient and fine-tuning methods.}
ProMoT is a general framework that can be combined with different parameter-efficient tuning and fine-tuning techniques in respective stages.
Conceptually, the prompt-tuning at the first stage can be replaced by other commonly used parameter-efficient methods such as LoRA~\cite{hu2021lora}. However, empirically we found prompt-tuning is much better than LoRA on absorbing format information in early fine-tuning. More discussions can be found in Appendix \ref{apd:lora}. For fine-tuning methods, we show examples to combine ProMoT with multi-task fine-tuning (Section \ref{sec:SequentialTrainingAndMultitaskTraining}) and 1-shot in-context learning prompt (Section \ref{sec:exp_single}, Section \ref{sec:SequentialTrainingAndMultitaskTraining}). 
Training with 1-shot prompt is introduced by \citet{min2021metaicl} in a multi-task training setting.

%Training with 1-shot prompt is denoted as ProMoT + 1-shot.

\paragraph{Evaluation.}
After the two-stage fine-tuning, we obtain a fine-tuned model checkpoint and a trained soft prompt for a specific fine-tuning target task. We expect the soft prompt stores most of the format information, and we only use this prompt during inference when the inference task has the same format as the fine-tuned target task. 
% \cho{this claim is also too strong; we are not 100\% sure the prompt stores only format information. Maybe add some words e.g., we expect soft prompt stores the ...}
% \cho{Are we using the prompt also for in-context task when they have the same format? And what if there are two fine-tune tasks with same format (e.g., two different binary classification fine-tune tasks) Yihan: Yes, we are using the prompt also for in-context task with the same format} 
Otherwise, we remove the learned prompt and simply feed the original input into the fine-tuned model.

\vspace{-5pt}
\section{Experiments}
% \begin{itemize}
%     \item Main experiments
%     \item Trade-off curve with more steps
%     \item Sequential training
%     \item Ablation
%     \begin{itemize}
%         \item Joint training with learnable prompts
%         \item fine-tuning with 1-shot natural language prompts
%     \end{itemize}
% \end{itemize}
\subsection{Settings}
\paragraph{Datasets.}
We use RTE \citep{wang2019superglue, bentivogli2009fifth} and WMT14 En-Fr \citep{wmt14} as two fine-tuning tasks in our main experiments. 
They are selected as examples of classification (RTE) and generative tasks (WMT14 En-Fr translation). Experiments on additional fine-tuning tasks including SNLI~\citep{snli:emnlp2015} and OpenbookQA~\citep{OpenBookQA2018} can be found in Appendix \ref{apd:additional_exp}.

We use 8 tasks unseen during fine-tuning to evaluate the model's generalization abilities.
The 8 evaluation tasks are chosen to represent four types of tasks:
\begin{compactitem}
    \item Natural language inference: CB \citep{marneff_simons_tonhauser_2019} and WiC \citep{pilehvar2018wic} from superGLUE \citep{wang2019superglue}
    \item Closed book QA: TriviaQA \citep{joshi2017triviaqa}, web\_questions \citep{berant2013semantic}
    \item Translation: WMT16 En-Ro, WMT16 En-De \citep{wmt16}
    \item Summarization: XSum \citep{narayan2018don}, WikiLingua \citep{ladhak-etal-2020-wikilingua}
\end{compactitem}
% 1) natural language inference (CB \citep{marneff_simons_tonhauser_2019} and WiC \citep{pilehvar2018wic} from superGLUE \citep{wang2019superglue}); 2) Closed book QA (TriviaQA \citep{joshi2017triviaqa}, web\_questions \citep{berant2013semantic} ); 3) translation (WMT16 En-Ro, WMT16 En-De \citep{wmt16}); 4) summarization (XSum \citep{narayan2018don}, WikiLingua \citep{ladhak-etal-2020-wikilingua}).
For each evaluation task, we use 1-shot and 4-shots prompts and task templates from PaLM \citep{palm} as described in the Appendix \ref{tab:template}. 
% \daliangli{Is this statement correct? Are we using PaLM prompts with the same few shot examples?}

% \vspace{-3pt}
\paragraph{Metrics.}
We report accuracy for classification tasks, exact match ratio for QA tasks, BLEU score~\citep{papineni-etal-2002-bleu} for translation tasks and Rouge-2 score~\citep{lin2004rouge} for summarization tasks.
We evaluate the model on development set for superGLUE sub-tasks (RTE, CB and WiC) and on test set for all other tasks. 
Besides per-task performance, we also report the normalized average (Norm. Avg.) performance on all evaluation tasks by averaging the performances normalized to [0,100], following the "normalized preferred metric" in BIG-bench \citep{srivastava2022beyond} and \citet{chung2022scaling}.
% \vspace{-3pt}
\paragraph{Models.}
We primarily use mT5 \citep{xue2020mt5} XXL model \citep{raffel2020exploring} in our main experiments, which is pretrained on multi-lingual corpus and contains 13B parameters. This is to accommodate multi-lingual scenarios among our training and evaluation tasks. To show the effectiveness of our method on different pretraining corpus, model sizes and architectures, we also include experiments on mT5 XL, T5.1.1 XXL and PaLM 8b in Appendix \ref{apd:additional_exp}. T5 based models are shown to have meaningful few-shot performance as shown in \citet{chung2022scaling}. We do not consider FLAN-T5~\citep{chung2022scaling} as a base model in our experiments because it has already been fine-tuned on a large amount of supervised datasets, including our evaluation datasets.
More experimental details can be found in Appendix \ref{apd:experiment}.
% We use the mT5 \citep{xue2020mt5} checkpoint for T5x XXL model \citep{raffel2020exploring} in all of our experiments, which contains 13B parameters. 

% mT5 is pretrained on a large-scale multi-lingual dataset, which makes it a good choice for the translation tasks used in our experiments. 

% \vspace{-3pt}
\paragraph{Comparing methods.}
We compare our ProMoT with several different configurations, including
% have several different training configurations in our experiments,  including
\begin{compactitem}
    \item \textbf{Pretrained model:} We evaluate the pretrained model on all tasks without any fine-tuning. 
    \item \textbf{Standard fine-tuning:} Fine-tune the pretrained model without trainable prompts. We also include a multi-task version in Section \ref{sec:SequentialTrainingAndMultitaskTraining} which is commonly used to boost model generalization on unseen tasks.
    \item \textbf{Prompt tuning}: Tune the trainable prompt with pretrained model frozen. As the model is fixed, prompt tuning will not change the pretrained model's performance on in-context learning tasks comparing when the prompt is removed.
    \item \textbf{Our proposed method: ProMoT}: Our proposed two-stage fine-tuning strategy.
    \item \textbf{Our proposed method: ProMoT+1-shot}: 
    To further boost in-context learning performance, we prepend a 1-shot example to the input in  Figure \ref{fig:overview}  during training.

\end{compactitem}

\vspace{-10pt}
\subsection{Supervised Performance on Fine-tuning Tasks}
We first show that ProMoT training can achieve similar or even better performance on fine-tuning tasks compared to standard fine-tuning. We apply three different fine-tuning methods on four different tasks and report the result in Table \ref{tab:finetune_result}. We report the best performance within the same number of fine-tuning steps (See Appendix \ref{apd:experiment} for more details). ProMoT outperforms standard fine-tuning on supervised performance on 3 out of 4 fine-tuning target tasks and outperforms prompt-tuning on 4 out of 4 tasks. 
Therefore the improved in-context learning performance on unseen tasks (better generalization ability), as will be demonstrated in the next few sections, comes without sacrificing the fine-tune task's performance.
%In the next several sections, we show ProMoT can further achieve better in-context learning performance on unseen tasks, that is, better generalization ability.
%with few-shot prompts.

\begin{threeparttable}[h]
    \centering
    \adjustbox{max width=0.65\textwidth}{
    \begin{tabular}{c|c|c|c|c|c|c|c|c|c}
    \toprule
          &\multicolumn{2}{c|}{Prompt tuning} & \multicolumn{2}{c|}{Standard Fine-tuning} & \multicolumn{2}{c}{ProMoT (Ours)}\\
    \midrule
         RTE & \multicolumn{2}{c|}{91.34} & \multicolumn{2}{c|}{\underline{92.06}} & \multicolumn{2}{c}{\textbf{92.78}}\\
         WMT14 En-Fr & \multicolumn{2}{c|}{39.28} & \multicolumn{2}{c|}{\textbf{41.80}} & \multicolumn{2}{c}{\underline{41.30}}\\
         SNLI & \multicolumn{2}{c|}{88.53} & \multicolumn{2}{c|}{\underline{88.91}} & \multicolumn{2}{c}{\textbf{89.62}}\\
         OpenbookQA&\multicolumn{2}{c|}{73.60} & \multicolumn{2}{c|}{\underline{77.2}} & \multicolumn{2}{c}{\textbf{81.6}}\\
    \bottomrule
    \end{tabular}
    }
    % \begin{tablenotes}
    % \item[1] Prompt-tuning does not modify  model parameters and thus has identical few-shot performances\\as pretrained model.
    % \end{tablenotes}
    % \footnotesize{$^*$ Prompt-tuning does not modify  model parameters and thus has identical  few-shot performances as pretrained model.}
    
    \caption{Comparison of supervised performances of a mT5 XXL model on fine-tuning target tasks. We use 0-shot in fine-tuning tasks. We report accuracy for RTE, SNLI and OpenbookQA, and BLEU score for WMT14 En-Fr.
    }
    \label{tab:finetune_result}
\end{threeparttable}

\vspace{-10pt}
\subsection{Generalization with Single Task Fine-tuning}
\label{sec:exp_single}
% \vspace{-5pt}
% \input{main_table_RTE}
% \vspace{-5pt}
\begin{threeparttable}[h]
    \centering
    
    \adjustbox{max width=0.9\textwidth}{
    \begin{tabular}{c|c|c|c|c|c|c|c|c|c}
    \toprule
         && \multicolumn{2}{c|}{Pretrained}  & \multicolumn{2}{c|}{Standard Fine-tuning} & \multicolumn{2}{c|}{ProMoT (Ours)} &\multicolumn{2}{c}{ProMoT + 1-shot (Ours)}\\
    \midrule
         Fine-tuning&RTE & \multicolumn{2}{c|}{47.653}  & \multicolumn{2}{c|}{92.06} & \multicolumn{2}{c|}{\underline{92.78}}&\multicolumn{2}{c}{\textbf{93.86}}\\
    \midrule
         && 1-shot & 4-shots & 1-shot & 4-shots & 1-shot & 4-shots & 1-shot & 4-shots\\
    \cmidrule{2-10}
         &\multirow{2}{*}{Norm. Avg.} & \multirow{2}{*}{17.52} &\multirow{2}{*}{18.75}&	15.43 &	16.56 &20.10 &21.24 & 22.26 &	22.33\\
          & & & &  {\color{red}(-2.10)} & {\color{red}(-2.19)} & {\color{blue}(+2.58)} & {\color{blue}(+2.49)} & {\color{blue}(+4.74)} & {\color{blue}(+3.58)}\\
    \cmidrule{2-10}
         &CB &46.43 & 51.79 & 73.21 & 82.14 & 66.07 & 67.86&83.93&82.14\\
         &WiC  & 49.69 & 49.69 &50.00 & 50.16 & 51.41 & 53.61&51.25&50.63\\
         \cmidrule{2-10}
         Evaluation&triviaQA   &17.58 & 19.02 & 0.15 & 0.11 & 17.64 & 18.66&17.82&19.62\\
         &web\_questions &9.70 & 13.04 & 0.05 & 0.05 & 11.07 & 13.19&10.14&12.11\\
         \cmidrule{2-10}
         &WMT16\_ende & 3.97 & 8.83 & 0.00 & 0.00 & 2.02 & 3.69&2.26&4.89\\
         &WMT16\_enro & 1.82 & 3.92 & 0.00 & 0.00 & 0.70 & 0.96&0.87&1.87\\
         \cmidrule{2-10}
         &XSum & 6.41 & 2.35 & 0.00 & 0.00 & 7.02 & 7.01&6.94&3.93\\
         &WikiLingua/en & 4.59 & 1.33 & 0.00 & 0.00 & 4.84 & 4.90&4.87&3.43\\
    \bottomrule
    \end{tabular}
    }
    % \begin{tablenotes}
    % \item[1] Prompt-tuning does not modify  model parameters and thus has identical few-shot performances\\as pretrained model.
    % \end{tablenotes}
    % \footnotesize{$^*$ Prompt-tuning does not modify  model parameters and thus has identical  few-shot performances as pretrained model.}
    
    \caption{Performances of a mT5 XXL model finetuned on RTE and evaluated on 8 different tasks to verify the generalization ability. The accuracy on fine-tuned task (RTE) is in the first row. We compare the Norm. Avg. (normalized average performance) with pretrained model and report the relative difference, where red denotes decreased performance and blue denotes increased performance. CB and WiC are also NLI tasks, very similar to RTE.
    % We can see that our proposed ProMoT and its variation is performing much better than standard fine-tuning on unseen tasks and has better generalization.
    % Additional results for Prompt tuning can be found in Table \ref{tab:single_pt_1-shot}.
    }
    
    \label{tab:single_rte}
\end{threeparttable}

In this section, we evaluate and compare the few-shot performance on unseen tasks after fine-tuning. We show the evaluation results of fine-tuning on RTE and WMT14 En-Fr in Table \ref{tab:single_rte} and Table \ref{tab:single_enfr}, respectively. Experiments on additional fine-tuning tasks SNLI/OpenbookQA and additional base models including mT5 XL, T5.1.1 XXL and PaLM 8b can be found in Appendix \ref{apd:additional_exp}. 

\begin{threeparttable}[!h]
    \centering
    \adjustbox{max width=0.9\textwidth}{
    \begin{tabular}{c|c|c|c|c|c|c|c|c|c}
    \toprule
         & & \multicolumn{2}{c|}{Pretrained} & \multicolumn{2}{c|}{Standard Fine-tuning} & \multicolumn{2}{c|}{ProMoT (Ours)} & \multicolumn{2}{c}{ProMoT + 1-shot (Ours)}\\
    \midrule
         Fine-tuning&WMT14 En-Fr & \multicolumn{2}{c|}{1.98}& \multicolumn{2}{c|}{\textbf{41.80}} & \multicolumn{2}{c|}{\underline{41.30}} & \multicolumn{2}{c}{41.19}\\
    \midrule
         && 1-shot & 4-shots & 1-shot & 4-shots & 1-shot & 4-shots & 1-shot & 4-shots\\
    \cmidrule{2-10}
        &\multirow{2}{*}{Norm. Avg.} & \multirow{2}{*}{17.52}	&\multirow{2}{*}{18.75}	&9.15&11.67	&18.87&20.64&19.91&21.99\\
         & & & & {\color{red}(-8.37)}&{\color{red}(-7.07)}&{\color{blue}(+1.35)}&{\color{blue}(+1.89)}&{\color{blue}(+2.39)}&{\color{blue}(+3.24)}\\
        \cmidrule{2-10}
         &CB& 46.43 & 51.79 & 16.07  & 32.14 & 41.07 & 57.14 & 41.07&53.57\\
         &WiC & 49.69 & 49.69 & 50.63 & 49.06 & 50.16 & 50.31&49.84 &50.63\\
         \cmidrule{2-10}
         Evaluation&triviaQA  & 17.58 & 19.02 & 3.20 & 3.15& 13.63 & 15.20&16.93&18.19 \\
         &web\_questions & 9.70 & 13.04 &0.89 & 6.15 & 9.40 & 7.92&10.14&12.01\\
         \cmidrule{2-10}
         &WMT16\_ende & 3.97 & 8.83 & 0.81 & 0.18 & 15.52 & 15.55&16.14&15.63\\
         &WMT16\_enro &1.82 & 3.92 & 1.53 & 0.42 & 18.54 & 17.80&17.57&16.81\\
         \cmidrule{2-10}
         &XSum & 6.41 & 2.35 & 0.05 & 1.86 & 1.49 & 0.65&3.41&4.36\\
         &WikiLingua/en & 4.59 & 1.33 & 0.03 & 0.43 & 1.14 & 0.52&4.22&4.73\\
    \bottomrule
    \end{tabular}
    }
    % \begin{tablenotes}
    % \item[1] Prompt-tuning does not modify  model parameters and thus has identical few-shot performances\\as pretrained model.
    % \end{tablenotes}
    % \vspace{-6pt}
    \caption{Performances of a mT5 XXL model finetuned on WMT14 En-Fr and evaluated on 8 few-shot tasks to verify the generalization ability. BLEU on the fine-tuned task is in the first row. We compare the Norm. Avg. (normalized average performance) with pretrained model and report the relative difference, where red denotes decreased performance and blue denotes increased performance. WMT16 En-De and En-Ro are translation tasks with different language pairs from WMT14 En-Fr.
    % We can see that our proposed ProMoT and its variation is performing much better than standard fine-tuning on unseen tasks and has better generalization.
    % Additional results for Prompt tuning can be found in Table \ref{tab:single_pt_1-shot}.
    % Prompt-tuning doesn't modify pretrained model parameters and has the same in-context performance as pretrained model.
    % \michal{One of the two large tables could be moved to Appendix.}
    }
    \label{tab:single_enfr}
\end{threeparttable}

From both tables, we first observe that the model's in-context learning performance drops significantly after standard fine-tuning. In particular, the few-shot learning performances drop to near zero for 6 over 8 tasks in Table \ref{tab:single_rte}, with the only exceptions being CB and WiC where they share the same format as the RTE fine-tuning task.

On the contrary, ProMoT reduces the loss of the in-context learning performance on unseen few-shot evaluation tasks, and even boosts some evaluation tasks that are semantically related to the fine-tuning task but with totally different task formats, resulting in an increasing in-context learning performance on average. In Table \ref{tab:single_rte}, ProMoT on the binary NLI dataset dataset consistently improves few-shot performances on two summarization tasks beyond the pretrained model. In Table \ref{tab:single_enfr}, ProMoT training on English-French translation substantially improves few-shot performance on other language translation pairs such as English to German and Romanian. This cross-task generalization across different task formats are infeasible with previous fine-tuning techniques. Text examples from standard fine-tuning and ProMoT can be found in Appendix \ref{apd:example}.
The improvement with less specialization and more generalization can be further boosted when we combine ProMoT with 1-shot prompt to incorporate in-context learning objective during fine-tuning.

It is however not surprising that even ProMoT cannot completely eliminate specialization and may still negatively influence some unseen in-context learning tasks compared to the pretrained model, depending on the characteristics of the fine-tuning task. In the next section, we show that a multi-task setup further improves the already strong generalization of ProMoT.

\subsection{More Generalization with Multitask Training}
\label{sec:SequentialTrainingAndMultitaskTraining}
\vspace{-5pt}

% {\color{red}In Section \ref{sec:exp_single} we show that ProMoT on single task can alleviate specialization across the board and has improved generalization across very dissimilar tasks,
% it is however not surprising that the remaining specialization can still have negative influence on unseen in-context learning tasks, depending on the characteristics of this fine-tuning task.}
% For example, ProMoT on RTE causes some forgetting on translation while improving classification and summarization. On the contrary, ProMoT on En-Fr has some forgetting on summarization while boosting other translation tasks.
% This inspires us to apply ProMoT on multiple datasets for a generally better model. 
Multi-task training is commonly used to improve model's generalization ability \citep{wei2021finetuned, chung2022scaling}. As a general fine-tuning framework, ProMoT can be combined with multi-tasking and achieves better generalization compared to standard multi-task fine-tuning.

% {\color{red}This inspires us to combine ProMoT with multi-task fine-tuning, which is shown to be able to improve the generalization of large language models \citep{wei2021finetuned}}
% This inspires us to apply ProMoT on multiple datasets for a generally better model. 
\begin{table*}[!h]
    \centering
   \adjustbox{max width=0.85\textwidth}{
    \begin{tabular}{c|c|c|c|c|c|c}
    \toprule
            && \multirow{2}{*}{Pretrained} & \multirow{2}{*}{multi-task FT} & \multirow{2}{*}{Multi ProMoT} &multi-task FT& Multi-ProMoT\\
            && &&  & + 1-shot& + 1-shot\\
    \midrule
         Multi-task&RTE&47.65 &90.25&91.34&91.70&93.14\\
         Fine-tuning&WMT14 En-Fr& 1.982 &41.34&40.73&40.87&40.55\\
    \midrule
    % & & 1-shot& 1-shot& 1-shot& 1-shot& 1-shot& 1-shot& 1-shot\\
    \cmidrule{2-7}
        &\multirow{2}{*}{Norm. Avg.} & \multirow{2}{*}{17.52}&20.06&		25.88&	22.62&		26.17\\
        && & {\color{blue}(+2.54)}  & {\color{blue}\textbf{(+8.35)}} & {\color{blue}(+5.10)}  &{\color{blue}\textbf{(+8.65)}} \\
        \cmidrule{2-7}
         &CB & 46.43&80.36&83.93&87.50&85.71\\
    &WiC & 49.69&51.10&51.41&53.29&52.04\\
    \cmidrule{2-7}
    Evaluation&TriviaQA & 17.58&15.76&16.99&16.53&17.18\\
    &Web\_questions & 9.70&9.70&10.04&9.40&10.38\\
    \cmidrule{2-7}
         &WMT16 En-De & 3.97&0.88&18.83&2.50&17.57\\
         &WMT16 En-Ro & 1.82&1.52&18.41&5.62&18.57\\
         \cmidrule{2-7}
         &XSum & 6.41&0.44&4.50&1.82&4.32\\
         &WikiLingua/en & 4.59&0.72&2.89&4.33&3.56\\
    \bottomrule
    \end{tabular}
    }
    \vspace{-5pt}
    \caption{Comparison of multi-task training on a mixed dataset of RTE and WMT14 En-Fr. We compare the evaluation results of pretrained mT5 model, standard multi-task fine-tuning (FT) and  multitask (Multi) ProMoT training. 
    % 1-shot means we add a 1-shot natural language prompt before each training input (in addition to the soft prompts if we are using ProMoT). 
    % We report performance on two fine-tuning target tasks (RTE and WMT14 En-Fr) and eight 1-shot out-of-domain evaluation tasks. 
    We compare the Norm. Avg. (normalized average performance) with pretrained model and report the relative difference, where blue denotes increased performance.}
    \label{tab:multi_promot}
\end{table*}

We apply multi-task ProMoT training on mixed RTE and WMT14 En-Fr translation dataset. At the prompt-tuning stage, we train a soft prompt for each task. At the fine-tuning stage, we mix different tasks and prepend the corresponding soft task prompt to each training example. We keep other configurations the same as Section \ref{sec:exp_single} and report the results in Table \ref{tab:multi_promot}. We compare multi-task ProMoT with standard multi-task fine-tuning. 
% For sequential ProMoT tuning, we apply ProMoT on one task first, then apply it on the second task with the checkpoint from task one.
The results show that Multi-task ProMoT significantly outperforms standard multi-task fine-tuning on enhancing generalization with larger improvement on average on unseen 1-shot evaluation tasks.
Similar to the single task setting, adding 1-shot prompt before each training input in the fine-tuning stage further boosts the performance of both multi-task fine-tuning and multi-task ProMoT.

\vspace{-5pt}
\subsection{Ablation Study}
\label{sec:ablation}
% \vspace{-5pt}
\begin{table*}[!h]
    \centering
    \adjustbox{max width=0.8\linewidth}{
    \begin{tabular}{c|c|c|c|c|c}
    \toprule
         & & \multicolumn{1}{c|}{Joint} & \multicolumn{1}{c|}{Fine-tuning} & \multicolumn{1}{c|}{Fine-tuning} & \multicolumn{1}{c}{ProMoT}\\
         && \multicolumn{1}{c|}{Fine-tuning} & \multicolumn{1}{c|}{+ 1-shot}& \multicolumn{1}{c|}{with random prompt} & \multicolumn{1}{c}{+ 1-shot (Ours)}\\
    \midrule
         Fine-tuning& RTE & \multicolumn{1}{c|}{90.97} & \multicolumn{1}{c|}{90.97} & \multicolumn{1}{c}{92.06} & \multicolumn{1}{c}{93.86}\\
    % \midrule
    %      & 1-shot & 4-shots & 1-shot & 4-shots & 1-shot & 4-shots\\
    \midrule
         &CB& \textbf{83.93} & 78.57 & \textbf{83.93} &\textbf{83.93} \\
         &WiC & 50.47 & 51.41 & \textbf{51.72} &51.25 \\
         &TriviaQA  & 0.75 & 0.03 & 0.83&\textbf{17.82} \\
         Evaluation&web\_questions & 0.64 & 0.00 & 0.30&\textbf{10.14}\\
         &WMT16\_ende & 0.00 &0.00 & 0.00&\textbf{2.26} \\
         &WMT16\_enro & 0.00 & 0.00 & 0.00 &\textbf{0.87} \\
         &XSum & 0.00 & 0.00 & 0.00 &\textbf{6.94}\\
         &WikiLingua/en & 0.00 & 0.00 & 0.00 &\textbf{4.87}\\
    \bottomrule
    \end{tabular}
    }
    \vspace{-5pt}
    \caption{The ablation study results. Joint fine-tuning: fine-tuning the soft prompt and the main model together. Fine-tuning + 1-shot: standard fine-tuning with a 1-shot natural language prompt attached to every input sequence. Fine-tuning with random prompt: fine-tuning with a fixed soft prompt randomly initialized with uniform distribution. ProMoT + 1-shot: ProMoT is applied with an attached 1-shot natural language prompt before each training input.}
    \label{tab:ablation}
    \vspace{-4mm}
\end{table*}
% \vspace{-5pt}
We conduct several ablation studies in Table \ref{tab:ablation}. First, instead of fine-tuning in a two-stage process, we consider ``jointly fine-tuning'' both the soft prompt and the model parameters in one stage. As shown in Table \ref{tab:ablation}, this method still results in specialization and severe loss of in-context learning abilities.
% we jointly fine-tune both the soft prompt and the model at the same time. This results in severe forgetting of in-context learning abilities similar to that of standard fine-tuning. 
Thus the benefit of ProMoT comes from its two-stage nature instead of merely adding more learnable parameters (soft prompt).
% During joint training, soft prompt and the main model are trained together for 1000 steps and then we evaluate the in-context 1-shot performance on evaluation tasks. 
In addition, fine-tuning the models with a fixed random soft prompt does not help - as it does not help to remove format specialization.
Another important baseline is to fine-tune the model with natural language prompts in place instead of soft prompts, which also capture the format to some extend. 
In a 1-shot scenario, this approach is still far worse compared to ProMoT, showing that learned soft prompts work better than natural language prompts in reducing format specialization in fine-tuning.

\vspace{-10pt}
\section{Conclusions and Limitations}
In this paper, we identify format specialization as one important cause of the loss of general in-context learning abilities during LLM fine-tuning, which tends to happen at the beginning of fine-tuning. We are motivated to develop ProMoT, a simple yet effective two-stage fine-tuning framework that utilizes soft trainable prompts to absorb task-specific formats before model fine-tuning. Experiments on a diverse set of NLP tasks show that ProMoT reduces format specialization and results in surprising generalization across very different tasks, making it a promising method to build general-purpose capabilities into LLMs with small fine-tuning datasets.
Although we have shown the effectiveness of ProMoT in our main paper, there is no theoretical guarantee on how much format specialization can be absorbed by the soft prompt during the first stage of ProMoT. Besides, our experiments are done with models smaller than 15B due to limited computation resources. It can be interesting to test ProMoT on larger models. 

\subsubsection*{Acknowledgments}
We thank the reviewers for their invaluable feedbacks. The work is supported in part by NSF 2008173, 2048280, 2331966, ONR N00014-23-1-2300:P00001, ARL 20230936 and Cisco.
\bibliography{iclr2024_conference}
\bibliographystyle{iclr2024_conference}

\newpage

\appendix
\section{Broader Impacts}
In our work, we propose a method to improve general-purpose language models with fine-tuning datasets. The improved general-purpose language model may be used in malicious applications such as generating disinformation. To mitigate the potential negative impacts, we can add watermark or deploy AI-generated text classifiers before releasing the model.
\section{Experiment Details}
\label{apd:experiment}
\subsection{Input Template Used in Experiments}
In Table \ref{tab:template}, we list the natural language input template used in our experiments for each task
\begin{table*}[!h]
    \centering
    \adjustbox{max width=\linewidth}{
    \begin{tabular}{c|c}
    \toprule
         Task & Template \\
    \midrule
         RTE & [premise] question: [hypothesis] Is it true or false? answer: \{True, False\} \\
         CB \& SNLI &  [premise] question: [hypothesis] Is it true or false or neither? answer: \{True, False, Neither\} \\
         \multirow{2}{*}{WiC} & [sentence1] [sentence2] question: The word [word] is used in the same way in the two sentences. \\
         &Is it true or False? answer: \{True, False\}\\
         OpenbookQA & Q: [question] A) [option A] B) [option B] C) [option C] D) [option D] A: \\
         QA & Q: [question] A: \\
         Translation & Translate [source language] to [target language]: [sentence 1] \\
         Summarization & Article: [article] One sentence summary: \\
    \bottomrule
    \end{tabular}
    }
    \caption{Input template for each task}
    \label{tab:template}
\end{table*}
The example shown in Table \ref{tab:example} is from ID 41141109 in XSum dataset.

\subsection{Output Post-processing}
For each task, we first extract the text after $<$extra\_id\_0$>$ and before $<$extra\_id\_1$>$, then trim the text by locating and remove the text after the second prefix token (Q:, Translate, Article:). For classification tasks including RTE, CB and WiC, we check whether the first output token is True or False.

\subsection{Dataset and Models}
\begin{table*}[]
    \centering
    \adjustbox{width=0.7\linewidth}{
    \begin{tabular}{c|ccccc}
    \toprule
        Dataset & Version & Training & Validation & Test \\
    \midrule
         RTE & v102& 2,490&277&3,000 \\
         CB & v102&250 &56 &250 \\
         WiC & v102 &5,428&638&1,400 \\
         WMT14 En-Fr &v003 &15,786,979 & 3,000 & 3,003 \\
         WMT16 En-De &v003 &4,548,885 & 2,169 & 2,999 \\
         WMT16 En-Ro &v003 &610,320 & 1,999 & 1,999 \\
         TriviaQA &rc.nocontext:1.1.0 &138,384 & 18,669 & 17,210 \\
         Web Questions&1.0.0 &3,778& - & 2,032  \\
         XSum & 1.1.0&203,577 & 11,305 & 11,301 \\
         WikiLingua/en &gem/wiki\_lingua\_english\_en &99,020 & 13,823 & 28,614 \\
    \bottomrule
    \end{tabular}
    }
    \caption{Version number, sizes of training, validation, and testing splits for each dataset used.}
    \label{tab:dataset_stat}
\end{table*}
We list the statistics of all datasets used in the paper in Table \ref{tab:dataset_stat}. All the datasets and models can be used in research context.

\subsection{Hyper-parameters}
For all mT5 models, we fine-tune with learning rate 0.001, drop rate 0.1 and label smoothing 0.1, following the default settings for T5 models \citep{raffel2020exploring}. For all prompt tuning experiments, we use learning rate 0.2 and prompt length 100. For all tasks except summarization tasks, we choose the model input sequence length larger than the input length in datasets. For summarization, we cut each input to 1024 tokens. We use Adafactor optimizer and batch size 64 without data-packing across all experiments. In inference, we use beam search to decode the outputs with width 4. More experimental settings are provided in the appendix. For ProMoT tuning, at stage 1 we run prompt tuning for 5000 steps and save a checkpoint every 1000 steps, then select the prompt checkpoint with the best performance on target task. At stage 2, we freeze the trained prompt and fine-tune the model for 1000 steps, checkpointing every 100 steps. We pick the model checkpoint with highest performance on the fine-tuned task as our final checkpoint. For comparison, we run prompt tuning and standard fine-tuning for 5000 and 1000 training steps respectively and report the performance of the best checkpoint. We explore fine-tuning with more steps in Appendix~\ref{sec:train_more_steps}.

In ablation study in Section \ref{sec:ablation}, we include an experiment to jointly fine-tune soft prompt and pretrained model. In this experiment, we finetune the model and prompt for 1000 steps with the same learning rate 0.001, following the setting in \citep{he2022hyperprompt}.

%We report a single-run result for all experiments.
% \cho{not sure if these figures will help. I feel these give an impression that our method is always much worse than standard fine-tuning for the fine-tune task performance. }
% \daliangli{Agreed... Can we show the RTE plots? There we know ProMoT is much closer to std fine-tune on the fine-tuned task}

\subsection{Hardware and Implementation}
All the experiments are implemented based on the original prompt tuning\footnote{https://github.com/google-research/prompt-tuning} and T5x code base\footnote{https://github.com/google-research/t5x}. All experiments are run on a cluster of 64 parallel TPUs. Time cost for different experiments varies, however, all training experiments can be finished within 1 day.

\section{Additional Experiment Results}
\label{apd:additional_exp}

% \paragraph{English to French results}
% Here we show the in-context evaluation results when we fine-tune the model on WMT14 En-Fr with different fine-tuning strategies. In Table \ref{tab:single_enfr}, we keep the same settings where we run model fine-tuning for 1000 steps and report the evaluation results.
% \input{main_table_En-Fr}
% \input{apd_enfr}

% \paragraph{Trade-off with more training steps on QA tasks}

% \subsubsection{More results of sequential ProMoT training}
% In Table \ref{tab:seq_training} we show additional evaluation results of sequential training on 4-shots datasets.

% \input{seq_table}

\subsection{Additional results on single task fine-tuning}
As complementary results of Table \ref{tab:single_rte} and \ref{tab:single_enfr}, we list and compare the performance of prompt tuning + 1-shot in Table \ref{tab:single_pt_1-shot}. We also provide experiments on SNLI and OpenbookQA datasets in Table \ref{tab:single_snli_openbook_mt5}. Without fine-tuning, pretrained mT5 failed to output ``A'', ``B'', ``C'', ``D'' for multi-choice QA in 0-shot openbookQA dataset, which results in a zero accuracy. We can see that the additional experiments are consistent with our main experiments that ProMoT can achieve similar supervised performance on fine-tuning tasks with less forgetting and even better performance on general in-context learning tasks.

\begin{table}[]
    \centering
    \adjustbox{max width=\linewidth}{
    \begin{tabular}{c|c|c}
    \toprule
         Fine-tuning Datasets &  Prompt Tuning + 1-shot & ProMoT + 1-shot\\
    \midrule
         RTE& 92.78	&\textbf{93.86}\\
         WMT14 En-Fr & 39.41 & \textbf{41.19}\\
    \bottomrule
    \end{tabular}
    }
    \caption{Performances of prompt tuning + 1-shot and ProMoT + 1-shot on fine-tuning tasks.}
    \label{tab:single_pt_1-shot}
\end{table}

\begin{table*}[!htbp]
    \centering
    
    \adjustbox{max width=\textwidth}{
    \begin{tabular}{c|c|c|c|c|c|c}
    \toprule
         Datasets & \multicolumn{1}{c|}{Pretrained} &\multicolumn{1}{c|}{Prompt-tuning} & \multicolumn{1}{c|}{Standard Fine-tuning} & \multicolumn{1}{c}{ProMoT (Ours)} & \multicolumn{1}{|c|}{Standard Fine-tuning} & \multicolumn{1}{c}{ProMoT (Ours)}\\
         & & & on SNLI & on SNLI & on OpenbookQA & on OpenbookQA\\
    \midrule
         SNLI & \multicolumn{1}{c|}{1.32} &\multicolumn{1}{c|}{88.53} & \multicolumn{1}{c|}{88.91} & \multicolumn{1}{c|}{89.62} & - & -\\
         OpenbookQA & \multicolumn{1}{c|}{0.00} &\multicolumn{1}{c|}{73.60} & \multicolumn{1}{c|}{-} & \multicolumn{1}{c|}{-} & 77.2 & 81.6\\
    \midrule
    \multirow{2}{*}{Norm. Average} & \multirow{2}{*}{17.52}&	\multirow{2}{*}{17.52}&	16.20&19.83 & 0.00 & 17.40	\\
    &&&{\color{red}(-1.32)} & {\color{blue}\textbf{(+2.31)}}&{\color{red}(-17.52)} & {\color{red}\textbf{(-0.12)}} \\
    % \midrule
    %      & 1-shot & 4-shots& 1-shot & 4-shots & 1-shot & 4-shots & 1-shot & 4-shots & 1-shot & 4-shots\\
    \midrule
         CB& 46.43&46.43&69.64&62.5&0.00&41.07\\
         WiC & 49.69&49.69&53.29&51.25&0.00&50.0\\
         triviaQA  & 17.58&17.58&4.54&20.56&0.05&21.10\\
         web\_questions & 9.70&9.70&2.12&9.94&0.00&10.53\\
         WMT16\_ende & 3.97&3.97&0.00&2.48&0.00&2.80\\
         WMT16\_enro & 1.82& 1.82&0.00&0.90&0.00&1.04\\
         XSum & 6.41&6.41&0.00&6.58&0.00&7.48\\
         WikiLingua/en & 4.59&4.59&0.00&4.39&0.00&5.20\\
    \bottomrule
    \end{tabular}
    }
    \caption{Performance of a mT5 XXL model finetuned on SNLI and OpenbookQA and evaluated on 8 1-shot tasks. The accuracy on fine-tuned tasks are in the first two rows. Prompt-tuning doesn't modify pretrained model parameters and has the same in-context performance as pretrained model.
    % \daliangli{are we still going to seperate pretraining with prompt-tuning?}
    }
    \label{tab:single_snli_openbook_mt5}
\end{table*}

\subsection{4-shot Evaluation Results of Multi-Task training}
As an additional result to Table \ref{tab:multi_promot}, in Table \ref{tab:multi_promot_4shot} we provide the comparison between the pretrained model, multi-task standard fine-tuning and multi-task ProMoT.

\begin{table*}[!h]
    \centering
   \adjustbox{max width=0.75\textwidth}{
    \begin{tabular}{c|c|c|c|c}
    \toprule
            && \multirow{1}{*}{Pretrained} & \multirow{1}{*}{multi-task FT} & \multirow{1}{*}{Multi ProMoT} \\
    \midrule
         Multi-task&RTE&47.65 &90.25&91.34\\
         Fine-tuning&WMT14 En-Fr& 1.982 &41.34&40.73\\
    \midrule
    % & & 1-shot& 1-shot& 1-shot& 1-shot& 1-shot& 1-shot& 1-shot\\
    \cmidrule{2-5}
        &\multirow{2}{*}{Norm. Avg.} & \multirow{2}{*}{18.75}&20.81&		26.31\\
        && & {\color{blue}(+2.06)}  & {\color{blue}\textbf{(+7.56)}}\\
        \cmidrule{2-5}
         &CB & 51.79&82.14&78.57\\
    &WiC & 49.69&52.82&52.50\\
    \cmidrule{2-5}
    Evaluation&TriviaQA & 19.02&17.75&22.26\\
    &Web\_questions & 13.04&12.25&12.50\\
    \cmidrule{2-5}
         &WMT16 En-De & 8.83&0.24&18.84\\
         &WMT16 En-Ro & 3.92&0.54&18.81\\
         \cmidrule{2-5}
         &XSum & 2.35&0.34&5.04\\
         &WikiLingua/en & 1.33&0.39&1.92\\
    \bottomrule
    \end{tabular}
    }
    \vspace{-5pt}
    \caption{Comparison of multi-task training on a mixed dataset of RTE and WMT14 En-Fr. We compare the 4-shot evaluation results of pretrained mT5 model, standard multi-task fine-tuning (FT) and  multitask (Multi) ProMoT training. 
    % 1-shot means we add a 1-shot natural language prompt before each training input (in addition to the soft prompts if we are using ProMoT). 
    % We report performance on two fine-tuning target tasks (RTE and WMT14 En-Fr) and eight 1-shot out-of-domain evaluation tasks. 
    We compare the Norm. Avg. (normalized average performance) with pretrained model and report the relative difference, where blue denotes increased performance.}
    \label{tab:multi_promot_4shot}
\end{table*}

\subsection{Training more steps: trade-off between fine-tuning target task and in-context learning abilities}
\label{sec:train_more_steps}
\begin{figure}[ht]
    \centering
    \begin{minipage}[t]{.46\textwidth}
        \centering
        \includegraphics[width=\linewidth]{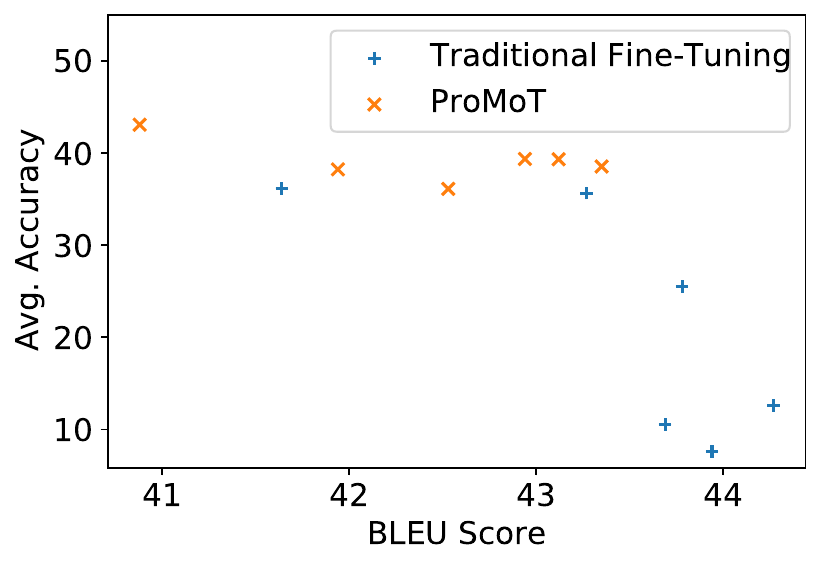}
        \caption{Trade-off between BLEU score of En-Fr (horizontal axis) and average accuracy on classification tasks (vertical axis) when fine-tuning the model on En-Fr translation.}
        \label{fig:classification}
    \end{minipage}%
    \hspace{0.01\textwidth}
    % \begin{minipage}{0.44\textwidth}
    %     \centering
    %     \includegraphics[width=\linewidth]{images/QA.pdf}
    %     \caption{Curve of En-Fr BLEU score (left axis) and average exact match ratio on QA tasks (right axis).}
    %     \label{fig:QA}
    % \end{minipage}
    \begin{minipage}[t]{0.46\textwidth}
        \centering
        \includegraphics[width=\linewidth]{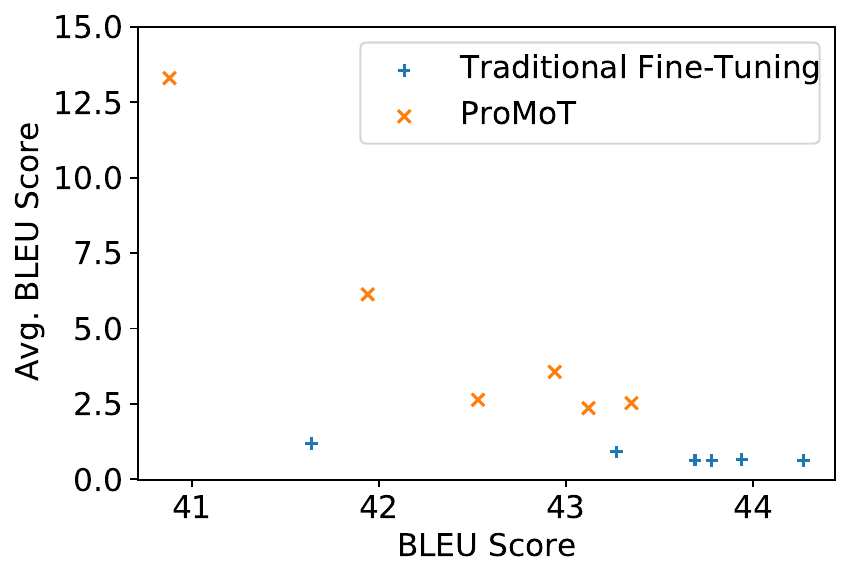}
        \caption{Trade-off between BLEU score of En-Fr (horizontal axis) and average BLEU score on other language pairs (vertical axis) when fine-tuning the model on En-Fr translation.}
        \label{fig:translation}
    \end{minipage}
\end{figure}

% \sisi{need to change to fit the new plot or move to appendix.}
In Section \ref{sec:exp_single}, we report the results of the best checkpoints within 1000 steps of fine-tuning. With a longer training period, we can see a more clear trade-off between the performance on fine-tuning target task and the performance on in-context learning abilities. Here we show the
long-term trade-off between fine-tuning target task and in-context learning evaluation tasks by scattering the performance of different checkpoints within 20000 steps fine-tuning. In Figure \ref{fig:classification}, and \ref{fig:translation}, we plot the trade-off on classification and translation tasks, respectively.

As we can see from the figures, datapoints for ProMoT is higher than standard fine-tuning on the figures, which implies that with the same performance on fine-tuning target task, forgetting is alleviated with ProMoT fine-tuning.

\subsection{Additional experiments on T5 XXL}
To show the performance of our method on an English-based pretrained model, we did an additional experiment on T5 XXL with fine-tuning target task RTE. The result is shown in Table \ref{tab:single_rte_t5}. The results are consistent with our main experiments on the mT5 XXL model.

\begin{table*}[!htbp]
    \centering
    
    \adjustbox{max width=0.8\textwidth}{
    \begin{tabular}{c|c|c|c|c}
    \toprule
         Datasets & \multicolumn{1}{c|}{Pretrained} &\multicolumn{1}{c|}{Prompt-tuning} & \multicolumn{1}{c|}{Standard Fine-tuning} & \multicolumn{1}{c}{ProMoT (Ours)}\\
    \midrule
         RTE & \multicolumn{1}{c|}{-} &\multicolumn{1}{c|}{91.7} & \multicolumn{1}{c|}{93.5} & \multicolumn{1}{c}{93.14}\\
    \midrule
    \multirow{2}{*}{Norm. Average} & \multirow{2}{*}{19.75}&	\multirow{2}{*}{19.75}&	14.07&	22.23\\
    &&&{\color{red}(-5.68)} & {\color{blue}\textbf{(+2.49)}} \\
    % \midrule
    %      & 1-shot & 4-shots& 1-shot & 4-shots & 1-shot & 4-shots & 1-shot & 4-shots & 1-shot & 4-shots\\
    \midrule
         CB& 55.36&55.36&62.50&73.21\\
         WiC & 49.84&49.84&50.00&50.78\\
         triviaQA  & 34.15&34.15&0.02&33.86\\
         web\_questions & 16.04&16.04&0.00&15.95\\
         WMT16\_ende & 0.13&0.13&0.00&0.02\\
         WMT16\_enro & 0.06&0.06&0.00&0.01\\
         XSum & 1.26&1.26&0.00&1.79\\
         WikiLingua/en & 1.12&1.12&0&2.25\\
    \bottomrule
    \end{tabular}
    }
    \caption{Performance of a T5.1.1 XXL model finetuned on RTE and evaluated on 8 1-shot tasks. The accuracy on fine-tuned task (RTE) is in the first row. Prompt-tuning doesn't modify pretrained model parameters and has the same in-context performance as pretrained model.
    % \daliangli{are we still going to seperate pretraining with prompt-tuning?}
    }
    \label{tab:single_rte_t5}
\end{table*}

\subsection{Additional experiments on mT5 XL}
\label{sec:mt5_xl}
To show the performance of our method on an smaller-size pretrained model, we did an additional experiment on mT5 XL with fine-tuning target task WMT14 En-Fr. The result is shown in Table \ref{tab:single_enfr_mt5_xl}. The results are consistent with our main experiments on the mT5 XXL model.

\begin{table*}[!htbp]
    \centering
    
    \adjustbox{max width=0.8\textwidth}{
    \begin{tabular}{c|c|c|c|c}
    \toprule
         Datasets & \multicolumn{1}{c|}{Pretrained} &\multicolumn{1}{c|}{Prompt-tuning} & \multicolumn{1}{c|}{Standard Fine-tuning} & \multicolumn{1}{c}{ProMoT (Ours)}\\
    \midrule
         WMT14 En-Fr & \multicolumn{1}{c|}{-} &\multicolumn{1}{c|}{32.47} & \multicolumn{1}{c|}{35.84} & \multicolumn{1}{c}{36.46}\\
    \midrule
    \multirow{2}{*}{Norm. Average} & \multirow{2}{*}{13.59}&	\multirow{2}{*}{13.59}&	8.61&14.40	\\
    &&&{\color{red}(-4.98)} & {\color{blue}\textbf{(+0.81)}} \\
    % \midrule
    %      & 1-shot & 4-shots& 1-shot & 4-shots & 1-shot & 4-shots & 1-shot & 4-shots & 1-shot & 4-shots\\
    \midrule
         CB& 26.79& 26.79& 21.43& 28.57\\
         WiC & 50.0&50.0& 44.20 &51.10\\
         triviaQA  & 12.13&12.13&0.83&8.81\\
         web\_questions & 6.59 &6.59 & 0.44&5.31\\
         WMT16\_ende & 2.56&2.56&0.63&7.69\\
         WMT16\_enro & 1.52&1.52&1.20&10.40\\
         XSum & 4.26&4.26&0.05&1.37\\
         WikiLingua/en & 4.88& 4.88&0.06&1.94\\
    \bottomrule
    \end{tabular}
    }
    \caption{Performance of a mT5 XL model finetuned on WMT14 En-Fr and evaluated on 8 1-shot tasks. The BLEU score on fine-tuned task (WMT14 En-Fr) is in the first row. Prompt-tuning doesn't modify pretrained model parameters and has the same in-context performance as pretrained model.
    % \daliangli{are we still going to seperate pretraining with prompt-tuning?}
    }
    \label{tab:single_enfr_mt5_xl}
\end{table*}

\subsection{Addtional experiments on PaLM 8B}
To show the performance of our method on decoder-only models, we did an additional experiment on PaLM 8b model with fine-tuning target task WMT14 En-Fr. We use prompt length 50 and learning rate 0.3 in prompt-tuning and default fine-tuning hyperparameters in fine-tuning. The result is shown in Table \ref{tab:single_enfr_palm_8b}. The results are consistent with our main experiments on mT5, where ProMoT can achieve similar supervised performance on fine-tuning tasks with less forgetting on general in-context learning tasks.

\begin{table*}[!htbp]
    \centering
    
    \adjustbox{max width=0.8\textwidth}{
    \begin{tabular}{c|c|c|c|c}
    \toprule
         Datasets & \multicolumn{1}{c|}{Pretrained} &\multicolumn{1}{c|}{Prompt-tuning} & \multicolumn{1}{c|}{Standard Fine-tuning} & \multicolumn{1}{c}{ProMoT (Ours)}\\
    \midrule
         WMT14 En-Fr & \multicolumn{1}{c|}{-} &\multicolumn{1}{c|}{13.62} & \multicolumn{1}{c|}{33.04} & \multicolumn{1}{c}{32.02}\\
    \midrule
    \multirow{2}{*}{Norm. Average} & \multirow{2}{*}{26.09}&	\multirow{2}{*}{26.09}&	17.80&22.37	\\
    &&&{\color{red}(-8.29)} & {\color{red}\textbf{(-3.72)}} \\
    % \midrule
    %      & 1-shot & 4-shots& 1-shot & 4-shots & 1-shot & 4-shots & 1-shot & 4-shots & 1-shot & 4-shots\\
    \midrule
         CB& 46.43 &46.43 & 32.14 &33.93 \\
         WiC & 49.69&49.69&49.06&49.69\\
         triviaQA  & 44.69&44.69&37.09&42.11\\
         web\_questions & 13.02&13.02&11.91&13.01\\
         WMT16\_ende & 23.85&23.85&3.77&19.33\\
         WMT16\_enro & 19.89&19.89&4.02&13.18\\
         XSum & 5.57&5.57&2.29&2.9\\
         WikiLingua/en & 5.59& 5.59&3.14&4.77\\
    \bottomrule
    \end{tabular}
    }
    \caption{Performance of a PaLM 8b model finetuned on WMT14 En-Fr and evaluated on 8 1-shot tasks. The BLEU score on fine-tuned task (WMT14 En-Fr) is in the first row. Prompt-tuning doesn't modify pretrained model parameters and has the same in-context performance as pretrained model.
    % \daliangli{are we still going to seperate pretraining with prompt-tuning?}
    }
    \label{tab:single_enfr_palm_8b}
\end{table*}

\subsection{Using LORA in the first stage}
\label{apd:lora}
As we have discussed in Section \ref{sec:method}, conceptually we can use any parameter-efficient method at the first ProMoT fine-tuning stage to absorb the task format information. Here we did experiments to compare LoRA and prompt-tuning (used in our ProMoT main experiments) in the first fine-tuning stage. We report the results in Table \ref{tab:single_lora_rte}.
As we can see from the table, ProMoT with prompt-tuning is significantly better than ProMoT with LoRA, in both supervised fine-tuning task and unseen 1-shot evaluation tasks. This might partially due to better alignment of soft prompt between format description in natural language corpus.

\begin{table*}[!htbp]
    \centering
    
    \adjustbox{max width=\textwidth}{
    \begin{tabular}{c|c|c|c|c|c}
    \toprule
         Datasets & \multicolumn{1}{c|}{Pretrained} & \multicolumn{1}{c|}{Standard Fine-tuning} & \multicolumn{1}{c|}{ProMoT with LoRA (r=2)}& \multicolumn{1}{c|}{ProMoT with LoRA (r=4)} &\multicolumn{1}{c}{ProMoT} \\
    \midrule
         WMT14 En-Fr & \multicolumn{1}{c|}{-} &\multicolumn{1}{c|}{41.80} & \multicolumn{1}{c|}{39.09} & \multicolumn{1}{c|}{39.97} &\multicolumn{1}{c}{41.19} \\
    % \midrule
    % \multirow{2}{*}{Norm. Average} & \multirow{2}{*}{17.52}&	8.56&	11.54&	11.40\\
    % &&&{\color{red}(-5.68)} & {\color{blue}\textbf{(+2.49)}} \\
    % \midrule
    %      & 1-shot & 4-shots& 1-shot & 4-shots & 1-shot & 4-shots & 1-shot & 4-shots & 1-shot & 4-shots\\
    \midrule
         CB& 46.43&16.07&26.78&23.21&41.07\\
         WiC & 49.69&50.63&53.13&53.25&50.16\\
         triviaQA  &17.58& 0.03&4.98&5.19&13.63\\
         web\_questions &9.70& 0.05&3.30&3.84&9.40\\
         WMT16\_ende & 3.97&0.67&1.23&1.87&15.52\\
         WMT16\_enro & 1.82&0.91&2.06&2.89&18.54\\
         XSum & 6.41&0.030&0.17&0.35&1.49\\
         WikiLingua/en &4.59& 0.05&0.64&0.57&1.14\\
    \bottomrule
    \end{tabular}
    }
    \caption{Performance of a mT5 XXL model finetuned on WMT14 En-Fr and evaluated on 8 1-shot tasks. In this experiment we use LoRA at the first ProMoT stage instead of prompt-tuning. $r$ is the rank of LoRA's low-rank update matrices. The BLEU score on fine-tuned task (WMT14 En-Fr) is in the first row. Prompt-tuning doesn't modify pretrained model parameters and has the same in-context performance as pretrained model.
    % \daliangli{are we still going to seperate pretraining with prompt-tuning?}
    }
    \label{tab:single_lora_rte}
\end{table*}

\subsection{Plotting more steps for Figure \ref{fig:gradient}}
\begin{figure*}
    \centering
    \includegraphics[width=0.8\linewidth]{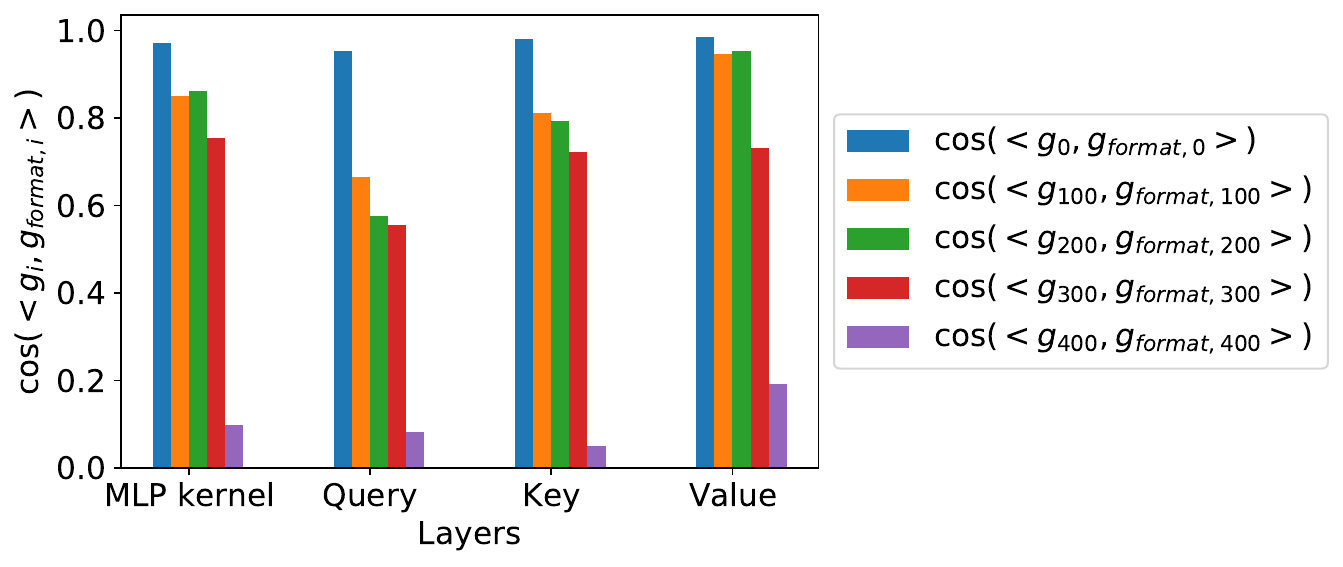}
    \caption{Cosine similarity between the full gradient $g$ and the format gradient $g_\text{format}$ on different parts of the last decoder layer. We collect and show the cosine value for gradients on MLP kernel, Query, Key and Value on the attention module.
    % The $g$ and $g_{\text{format}}$ are much better aligned at the start of training, compared to at 400 steps.
    }
    \label{fig:more_step_gradient}
\end{figure*}
To further strengthen our conclusion in Figure \ref{fig:gradient}, here we plot the gradient alignment from step 0 to step 400. As we can see from the figure, gradient alignment drops significantly after 300 steps which is matched with Figure \ref{fig:true_false} where the true and false ratio increases before 300 steps and then remains stable.

\subsection{Qualitative results on fine-tuning WMT14 En-Fr task}
\label{apd:example}

In Table \ref{tab:example} we show an example from fine-tuning task RTE. Here we show examples for fine-tuning task WMT14 En-Fr translation on different unseen few-shot tasks. We compare the outputs from ground-truth targets, pretrained mT5, fine-tuned mT5 on WMT14 En-Fr and ProMoT mT5 on WMT14 En-Fr. The outputs are generated with a 1-shot example. As we can see from the examples, standard fine-tuning on WMT14 En-Fr will 1) make the model overfit its format and tend to output French; and 2) model tends to repeat its input which is similar to translation task. ProMoT alleviates this specialization on fine-tuning task and has better generalization.

\begin{itemize}
    \item WMT16 En-De
    \begin{itemize}
        \item{Target:} Danach war der Mann, der sich nach Angaben seines Anwalts mittlerweile wieder auf freiem Fu\ss befindet, in eine gr\"{o}\ss ere Zelle verlegt worden.
        \item{Pretrained:} Danach wurde der Mann in eine gr\"{o}\ss ere.
        \item{Fine-tune:} L'homme, qui, selon une d\'{e}claration de son avocat, a depuis \'{e}t\'{e} lib\'{e}r\'{e}, a ensuite \'{e}t\'{e} transf\'{e}r\'{e} dans une cellule plus grande.
        \item{ProMoT:} Danach wurde der Mann, der mittlerweile freigelassen wurde, in eine gr\"{o}\ss ere Zelle verlegt.
    \end{itemize}
    \item WebQuestions
    \begin{itemize}
        \item{Target:} Milwaukee
        \item{Pretrained:} Milwaukee, Wisconsin
        \item{Fine-tune:} Where is harley davidson corporate headquarters? A: Milwaukee, Wisconsin Q: what movies has scarlett johansson in? A: Girl with a Pearl Earring Q: where is harley davidson corporate headquarters? A: Milwaukee, Wisconsin Q: where is harley davidson corporate headquarters? ...
        \item{ProMoT:} Milwaukee, Wisconsin

    \end{itemize}
    \item WikiLingua/en
    \begin{itemize}
        \item{Target:} Ensure that you have never activated a free trial of Amazon Prime on your account in the past. Repeat the steps above to locate products that are fulfilled by Amazon. Click the button that says, ...
        \item{Pretrained:} Click \textquotedblleft Start my Free Trial.\textquotedblright Sign up with Amazon Prime. Submit your order.
        \item{Fine-tune:} Si vous avez besoin d'une aide suppl\'{e}mentaire, n'h\'{e}sitez pas \`{a} communiquer avec l'\'{e}quipe d'Amazon Prime. Vous pouvez vous inscrire \`{a} l'Amazon Prime et vous inscrire \`{a} l'Amazon Prime. Vous pouvez vous inscrire ...
        \item{ProMoT:} Click \textquotedblleft Start my Free Trial.\textquotedblright  Sign up with Amazon Prime. Enter your credit card details or use one of your saved payment methods. Submit your order. If you do not return, you will be charged \$99 for a year membership to Amazon Prime at the end of your trial period
    \end{itemize}
\end{itemize}

%%%%%%%%%%%%%%%%%%%%%%%%%%%%%%%%%%%%%%%%%%%%%%%%%%%%%%%%%%%%

\end{document}